\documentclass[conference]{IEEEtran}
\IEEEoverridecommandlockouts
\usepackage{cite}
\usepackage{amsmath,amssymb,amsfonts}
\usepackage{algorithmic}
\usepackage{graphicx}
\usepackage{textcomp}
\usepackage[table]{xcolor}
\usepackage{subfigure}
\usepackage{caption}
\usepackage{subcaption}  
\def\BibTeX{{\rm B\kern-.05em{\sc i\kern-.025em b}\kern-.08em
    T\kern-.1667em\lower.7ex\hbox{E}\kern-.125emX}}
\begin{document}

\title{Fully Spiking Actor-Critic Neural Network for Robotic Manipulation\\

}

\author{\IEEEauthorblockN{1\textsuperscript{st} Liwen Zhang}
\IEEEauthorblockA{\textit{School of Astronautics} \\
\textit{Harbin Institute of Technology}\\
Harbin, China \\
liwenzhang@stu.hit.edu.cn}
\and
\IEEEauthorblockN{2\textsuperscript{nd} Heng Deng}
\IEEEauthorblockA{\textit{School of Astronautics} \\
\textit{Harbin Institute of Technology}\\
Harbin, China \\
dengh76@163.com}
\and
\IEEEauthorblockN{3\textsuperscript{rd} Guanghui Sun}
\IEEEauthorblockA{\textit{School of Astronautics} \\
\textit{Harbin Institute of Technology}\\
Harbin, China \\
guanghuisun@hit.edu.cn}
}

\maketitle

\begin{abstract}
This study proposes a hybrid curriculum reinforcement learning (CRL) framework based on a fully spiking neural network (SNN) for 9-degree-of-freedom robotic arms performing target reaching and grasping tasks. To reduce network complexity and inference latency, the SNN architecture is simplified to include only an input and an output layer, which shows strong potential for resource-constrained environments. Building on the advantages of SNNs—high inference speed, low energy consumption, and spike-based biological plausibility—a temporal progress-partitioned curriculum strategy is integrated with the Proximal Policy Optimization (PPO) algorithm. Meanwhile, an energy consumption modeling framework is introduced to quantitatively compare the theoretical energy consumption between SNNs and conventional Artificial Neural Networks (ANNs). A dynamic two-stage reward adjustment mechanism and optimized observation space further improve learning efficiency and policy accuracy. Experiments on the Isaac Gym simulation platform demonstrate that the proposed method achieves superior performance under realistic physical constraints. Comparative evaluations with conventional PPO and ANN baselines validate the scalability and energy efficiency of the proposed approach in dynamic robotic manipulation tasks.
\end{abstract}


\section{Introduction}
As computational efficiency becomes increasingly critical, a new generation of artificial neural networks (ANNs) is emerging. Among the most representative are Spiking Neural Networks (SNNs). Compared with traditional ANNs, SNNs offer several distinct advantages: communication based on sparse spike signals, rapid inference, event-driven computation, and significantly reduced energy consumption~\cite{tang2020reinforcement, yang2020recent}. In recent years, SNNs have found growing applications across diverse areas, including high-speed signal processing~\cite{bing2018survey}, object detection~\cite{zhou2020deep, qu2024spike, zanatta2023directly, liu2021spiking}, and image classification~\cite{mozafari2019bio, bu2023optimal, rathi2023exploring, yao2023attention}. These practical examples demonstrate the broad applicability and considerable potential of SNNs in addressing complex computational tasks under stringent energy constraints. 

In recent years, SNNs have attracted considerable attention and achieved promising results across a range of applications. Nevertheless, the deployment of SNNs in complex robotic control tasks has been relatively limited to date. A central difficulty lies in the non-differentiability of spiking activation functions, which prevents the direct application of conventional backpropagation techniques. Such non-differentiability introduces a fundamental limitation when precise temporal dynamics must be captured. To address this issue, researchers have introduced surrogate gradient methods. One widely used approach is the spatio-temporal backpropagation (STBP) algorithm~\cite{wu2018spatio}, which replaces the discontinuous spike function with a smooth, differentiable approximation. Consequently, gradient-based learning becomes applicable, which significantly facilitates the training process of SNNs. These developments have played a key role in advancing the integration of SNNs with reinforcement learning (RL), particularly for tasks involving dynamic control. As a result, there has been growing interest in exploring such combinations for real-time, energy-efficient decision-making in robotic systems~\cite{shrestha2018slayer, tang2020reinforcement, tang2021deep, oikonomou2022framework, park2025designing, li2023cbmc, jiang2025fully}.

While RL has been combined with SNNs in several studies, research on fully spiking actor-critic architectures for high-dimensional control tasks remains limited. In particular, the application of SNNs within RL frameworks to address complex, multi-stage tasks has not been systematically explored. This study explores the integration of curriculum reinforcement learning (CRL) into SNNs. The goal is to enable efficient control of robotic arms in multi-stage manipulation tasks.

The main contributions of this paper can be summarized as follows:

A hierarchical RL framework is proposed based on the SpikeGym~\cite{zanatta2024exploring} platform, integrating a progressive training strategy with a dynamic reward modulation mechanism. This framework significantly enhances the learning capacity of spiking neural agents in modeling complex behaviors and executing goal-directed tasks. It enables precise control of multi-degree-of-freedom robotic manipulators in high-dimensional operational scenarios. By structuring the learning process into distinct stages, the framework guides the policy to evolve gradually from simple to complex behaviors. This staged design not only alleviates the training difficulties caused by sparse reward signals but also improves the overall stability and sample efficiency of the system. 

To enable fair energy-aware comparison, a unified analytical scheme is introduced for estimating the consumption of SNNs and ANNs, integrating both neuronal activation rate and computation. By jointly accounting for these factors, the framework enables a fair and consistent comparison across different neural paradigms. Empirical results on the Isaac Gym~\cite{makoviychuk2021isaac} platform demonstrate that shallow SNN-based agents outperform ANN baselines with the same network depth in task performance, while achieving over an order of magnitude reduction in inference energy consumption. These findings validate the effectiveness of the proposed method and further highlight the promise of SNNs for energy-efficient control in robotic applications. 

\section{Methodology}

\subsection{SNN}

\begin{figure}[htbp]
    \centering
    \includegraphics[width=1\linewidth]{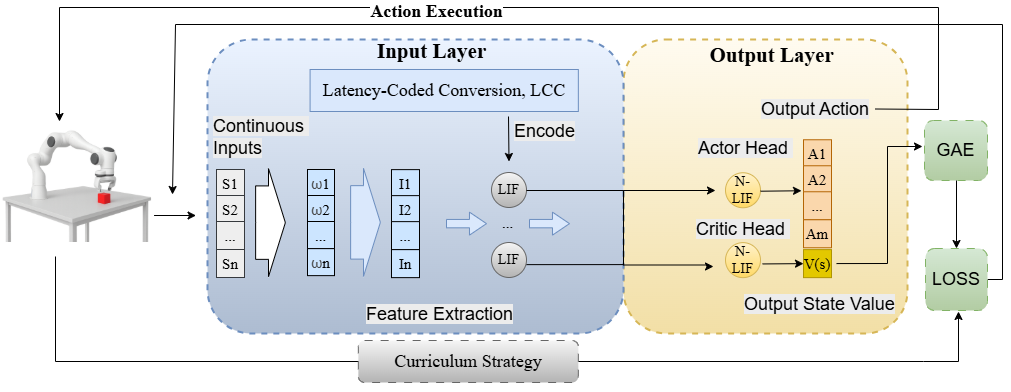}
    \caption{Curriculum-guided hybrid PPO architecture design.}
    \label{figureone}
\end{figure}

This study proposes a spiking neural control architecture for continuous action tasks, drawing inspiration from SpikeGym~\cite{zanatta2024exploring} but introducing key architectural modifications, as shown in Fig.~\ref{figureone}. The model employs Leaky Integrate-and-Fire (LIF) neurons in the input layer to directly convert continuous environmental observations into spike signals. Sensor values are treated as input currents and modulated by trainable weights, which are updated during learning to minimize task-specific loss. Spikes are generated intrinsically within the input layer, thereby eliminating the need for an explicit encoding stage. These spike signals are then passed to an output layer composed of non-spiking Leaky Integrate-and-Fire (N-LIF) neurons. The output neurons produce continuous signals for action selection and value estimation, based on their membrane potentials. The design enables a seamless mapping from continuous inputs to continuous outputs, with spike-based dynamics as the intermediate computation. The system is trained using surrogate gradient methods and a PPO framework with CRL, guided by dynamic reward feedback from the environment.

\subsection{Neuron Models}
LIF model~\cite{stein1965theoretical} is among the most commonly used spiking neuron models, owing to its balance between simplicity and computational efficiency. The LIF model is described by the following equation:

\begin{equation}\label{functionone}
\frac{dv_t}{dt} = \lambda \cdot (-v_t + \rho_t),
\end{equation}
where  $\lambda = \frac{1}{\tau},\ \rho_t = R \cdot I_t.$ $\tau$ denotes the membrane time constant, $v_t$ represents the membrane potential at time $t$, $R$ is the membrane resistance, and $I_t$ corresponds to the input current generated by pre-synaptic activity.

The output layer is designed using non-spiking LIF neurons, which do not emit discrete binary spikes. Instead, they continuously integrate synaptic input currents, leading to the evolution of their membrane potentials over time. To prevent the membrane potential from growing without bound, a hard-reset mechanism is employed. 

\subsection{Spike Encoding}

Continuous observations are transformed into spike-based signals using the LCC method. Each input feature fires a spike at a time step determined by its normalized magnitude. Larger values correspond to earlier spikes.

Suppose the observation comprises multiple feature vectors, which are combined and subsequently normalized to \([0, 1]^d\) using min-max scaling. Given a total of \(T\) simulation steps, the spike time for each feature is calculated as

\begin{equation}
    t_i = \left\lfloor (1 - \hat{x}_i) \cdot (T - 1) \right\rfloor,
\end{equation}
where \(\hat{x}_i\) denotes the normalized value of the \(i\)-th feature, and \(R \in \{0, 1\}^{d \times T}\) is a binary spike tensor in which each row contains a single spike at time step \(t_i\).

\subsection{Training Method}

For the robotic arm grasping task, this study adopts a unified RL framework that integrates curriculum learning with dynamic reward shaping. A model-free policy gradient method based on PPO is employed to ensure stable updates and sample efficiency. Training is guided by a time-varying composite dense reward function, enabling the progressive acquisition of manipulation skills—from coarse spatial alignment to precise positioning and stable grasping. 

At timestep $t$, the total reward $r_t$ is carefully constructed as a weighted sum of multiple functional components. The underlying reward structure aims to comprehensively guide the agent toward accomplishing complex manipulation objectives.

\begin{equation}\label{functionten}
r_t = r_{\text{prox-align}} + r_{\text{grip-geom}} + r_{\text{task}} + r_{\text{pose}} + p_{\text{pose}},
\end{equation}
where $r_{\text{prox-align}}$ emphasizes spatial proximity and precise alignment, 
$r_{\text{grip-geom}}$ quantifies the geometric symmetry of the gripper and the optimization of the grip gap,
$r_{\text{task}}$ provides critical incentives for successful task execution, 
$r_{\text{pose}}$ encourages the robotic arm to maintain a vertical and stable posture,
and $p_{\text{pose}}$ acts as a penalty term to suppress misalignment and instability.

Each component corresponds to a specific behavioral objective. The associated sub-rewards are described below.

\subsubsection{Proximity and Alignment Reward}

The alignment objective is designed to minimize the relative distance and planar displacement between the gripper and the cube, thereby ensuring accurate positioning and promoting robust manipulation performance.

\begin{align}\label{functioneleven}
r_{\text{prox-align}} &= \alpha_1 \left(1 - \tanh\left( \frac{\kappa_1(d + d_{\text{lf}} + d_{\text{rf}})}{3} \right)\right) \notag \\
&\quad + \alpha_2 \left(1 - \tanh(\kappa_2 \cdot d_{\text{align}})\right) \notag \\
&\quad + \alpha_3 \left(1 - \tanh(\kappa_3 \cdot d_{\text{mid}})\right) \notag \\
&\quad + \alpha_4 \left(1 - \tanh(\kappa_4 (\Delta x + \Delta y))\right),
\end{align}
where \(d\), \(d_{\mathrm{lf}}\), and \(d_{\mathrm{rf}}\) denote the distances from the cube center to the gripper midpoint, left and right finger, respectively. \(d_{\mathrm{align}}\) and \(d_{\mathrm{mid}}\) measure alignment errors in the XY plane and in 3D space. \(\Delta x\) and \(\Delta y\) represent offsets along the \(x\) and \(y\) axes. The coefficients \(\kappa_1\), \(\kappa_2\), \(\kappa_3\), and \(\kappa_4\) serve as scaling factors that adjust the reward’s sensitivity to these errors.

\subsubsection{Gripper Geometry Reward}

The gripper-related reward component promotes finger symmetry and appropriate grip width to improve grasp stability and mechanical robustness.

\begin{align}\label{functiontwelve}
r_{\text{grip-geom}} &= 
\beta_1 \left(1 - \tanh(\xi_1 \cdot |z_{\text{lf}} - z_{\text{rf}}|)\right) \notag \\
&\quad + \beta_2 \left(1 - \tanh(\xi_2 \cdot |z_{\text{mid}} - z_{\text{cube}}|)\right) \notag \\
&\quad + \beta_3 \cdot \exp(-\xi_3 \cdot |g - g^\ast|),
\end{align}
where \( z_{\text{lf}} \) and \( z_{\text{rf}} \) measure the vertical difference between the left and right fingers,  
\( z_{\text{mid}} - z_{\text{cube}} \) indicates the vertical misalignment between the gripper midpoint and the cube,  
and \( g - g^\ast \) evaluates the deviation of the gripper’s opening width from the optimal value.  
The scale factors \( \xi_1 \), \( \xi_2 \), and \( \xi_3 \) control the sensitivity to finger symmetry, vertical misalignment, and width deviation, respectively.

\subsubsection{Task Completion Reward}

High-value rewards are issued upon successful grasping:

\begin{align}
r_{\text{task}} &=
\gamma \cdot \mathbf{1}_{\text{grasped}} \cdot \varsigma \notag,
\end{align}
where $\mathbf{1}_{\text{grasped}}$ indicates whether the object is successfully grasped. $\varsigma$ denotes the reward for a successful grasp.

\subsubsection{Orientation and Stability Reward}

The posture control component of the reward function encourages vertical orientation of the end-effector and penalizes planar misalignment.

\begin{equation}
r_{\text{pose}} = \delta_1 \cdot \left| \left( \mathcal{R}(q_{\text{eef}}) \cdot \mathbf{z} \right)_z \right| + \delta_2 \cdot \left(1 - \tanh(\vartheta_1 \cdot \varepsilon_{xy})\right)
\end{equation}
\begin{equation}
p_{\text{pose}} = \vartheta_2 \cdot \varepsilon_{xy} \cdot \lambda - \vartheta_3 \cdot (1 - \nu_z) \cdot \lambda
\end{equation}

where \( r_{\text{pose}} \) encourages an upright end-effector posture and penalizes planar misalignment. The term \( \left( \mathcal{R}(q_{\text{eef}}) \cdot \mathbf{z} \right)_z \) measures the verticality of the end-effector by projecting the rotated \( z \)-axis onto the global vertical direction. The variable \( \varepsilon_{xy} \) denotes the planar alignment error, \( \nu_z \) represents the verticality score, and \( \lambda \) is a scaling factor that modulates the magnitude of penalization. The penalty term \( p_{\text{pose}} \) suppresses unstable configurations by weighting both horizontal deviation and vertical misalignment. The shaping parameter \( \vartheta_1 \) controls the nonlinearity of the alignment reward, while \( \vartheta_2 \) and \( \vartheta_3 \) determine the contributions of planar and vertical penalties, respectively.

\begin{figure}[htbp]
    \centering
    \includegraphics[width=0.4\textwidth]{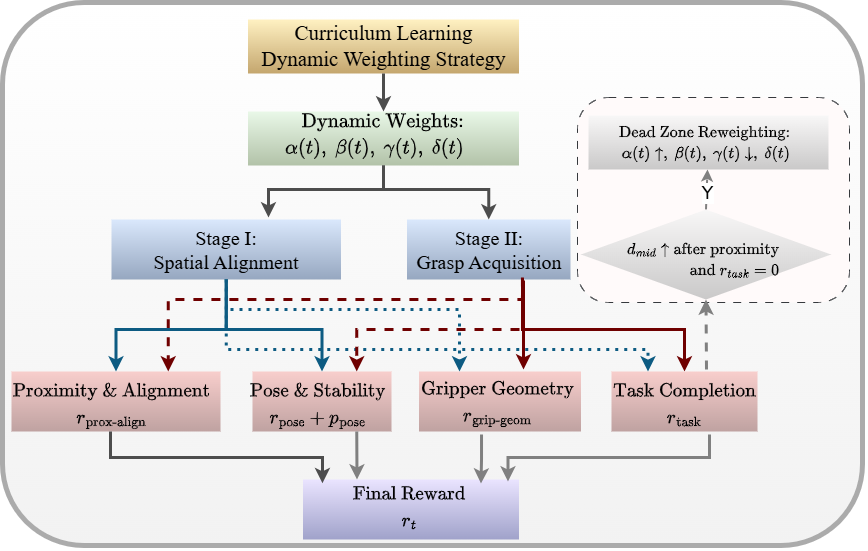}
    \caption{Dynamic reward weighting within a CRL framework is employed for staged robotic manipulation training. Solid arrows denote reward components prioritized at each stage, while dashed arrows indicate components maintained with reduced emphasis throughout training.}
    \label{figuretwo}
\end{figure}

To guide the agent toward key subgoals during training, a CRL approach is employed. Reward weights \(\alpha_i(t)\), \(\beta_i(t)\), \(\gamma(t)\), and \(\delta_i(t)\) vary dynamically with the global timestep \(t\), allowing adjustment of each reward’s influence at different stages. Figure~\ref{figuretwo} illustrates the curriculum structure and the evolution of reward weights over time. The training process is divided into two distinct stages:

\paragraph{\textbf{Stage I – Spatial Awareness} \ensuremath{(\mkern-2mu t\mkern-4mu <\mkern-4mu T_1\mkern-1mu)}}
The initial training stage is governed by a reward function that assigns primary weight to spatial proximity, alignment precision, and postural stability. This configuration is intended to facilitate agent-environment interaction while progressively shaping core coordination capabilities and control strategies.

\paragraph{\textbf{Stage II – Grasp Strategy Reinforcement (\(t \geq T_1\))}} 
At this stage, the reward function emphasizes key aspects of the grasping task, including gripper symmetry, finger-object spatial relationships, and their effect on the opening mechanism. It also promotes strategy refinement and adaptive response to object position variations. By reinforcing strategies closely linked to grasping, the reward guides the agent toward achieving precise and robust manipulation.

While curriculum-guided reward weighting structures the learning process in two stages, policy regressions may still occur in Stage~II due to local instability or noise. These regressions are particularly common when the agent has reached close proximity to the target but fails to complete a grasp. In such cases, the end-effector may drift, leading to reward inconsistency and unstable convergence.

To mitigate this issue, a dead zone-aware reweighting mechanism is proposed to dynamically adjust the reward structure in response to policy regressions. When the distance between the gripper and the target ($\mathit{d}_{\text{mid}}$) increases after a previous low value, and no grasp success is recorded ($r_\text{task}=0$), a dead zone is detected. The system then temporarily reshapes the reward structure by increasing the weights of alignment-related terms ($\alpha(t)$, $\beta(t)$) and decreasing the emphasis on grasp-specific terms ($\gamma(t)$, $\delta(t)$). This mechanism enables local policy correction without interrupting the curriculum schedule. It acts as a soft recovery layer, steering the agent back to meaningful transitions and reducing the risk of premature optimization.

\subsubsection{Estimation of Energy Consumption} \label{energy}
An analytical modeling approach is employed to compare the computational energy characteristics of SNNs and ANNs. This approach relies on operation-level energy profiling~\cite{rathi2021diet, rueckauer2017conversion}. Specifically, the encoder layer in SNNs typically employs floating-point multiply-accumulate (MAC) operations, whereas the subsequent fully connected layers predominantly utilize accumulate (AC) operations. In contrast, ANNs consistently perform MAC operations across all layers~\cite{jiang2025fully}.

Unlike conventional energy estimation methods that rely solely on operation counts, the proposed approach incorporates a detailed analysis of neuronal sparsity at each layer. Accordingly, the estimation of energy consumption becomes more accurate and structurally informative.

\paragraph{SNN Spike Rate and Energy Modeling}

The average spike rate of the input layer is defined as:
\begin{equation}
r = \frac{1}{B N_1} \sum_{b=1}^{B} \sum_{i=1}^{N_1} s_i^{(b)}
\end{equation}
where \(s_i^{(b)} \in \{0, 1\}\) indicates whether neuron \(i\) fired in sample \(b\).

For the output layer, the membrane activation rate is defined as:
\begin{equation}
r_{\text{mem}} = \frac{1}{N_2} \sum_{i=1}^{N_2} \mathbb{I}\left(\sum_{t=1}^{T}|V_i(t)|>0\right)
\end{equation}
where \(V_i(t)\) is the membrane potential of neuron \(i\) at time \(t\), and \(\mathbb{I}(\cdot)\) is the indicator function.

The estimated total energy consumption is:
\begin{align}\label{eq:snn_energy}
E_{\text{SNN}} = B T \Big[
& N_1 r (N_0 \alpha_m + (N_0{-}1) \alpha_a) \notag \\
& + N_2 r_{\text{mem}} N_1 \alpha_a
\Big]
\end{align}

\noindent where \(B\) is batch size, \(T\) is simulation steps, \(N_0\), \(N_1\), \(N_2\) are layer sizes, and \(\alpha_m\), \(\alpha_a\) are the energy costs of multiplication and addition (set to 4.6 and 0.9 pJ~\cite{kundu2021hire, yin2021accurate, yao2023attention, jiang2025fully}).

\paragraph{ANN Activation and Energy Modeling}

To ensure a fair comparison of energy consumption, the analytical model is also applied to ANNs.

The activation rates for the input and output layers are:
\begin{align}
r_{\text{in}} &= \frac{1}{B N_0} \sum_{b,i} \mathbb{I}(x_i^{(b)} > 0), &
r_{\text{out}} &= \frac{1}{B N_1} \sum_{b,j} \mathbb{I}(h_j^{(b)} > 0)
\end{align}

\noindent where \( x_i^{(b)} \) and \( h_j^{(b)} \) are activations of input and output neurons, respectively.

The estimated total energy is:
\begin{align}\label{eq:ann_energy}
E_{\text{ANN}} = B T \Big[ 
& N_1 r_{\text{in}} (N_0 \alpha_m + (N_0{-}1) \alpha_a) \notag \\
& + N_2 r_{\text{out}} (N_1 \alpha_m + (N_1{-}1) \alpha_a) \Big]
\end{align}

\noindent where \( B \) is batch size, \( T \) is simulation steps, \( N_0, N_1, N_2 \) are layer sizes, and \( \alpha_m, \alpha_a \) are energy costs of multiplication and addition.

\section{\label{rel} EXPERIMENTS}

\begin{figure}[htbp]
  \centering
{%
    \includegraphics[width=0.48\linewidth]{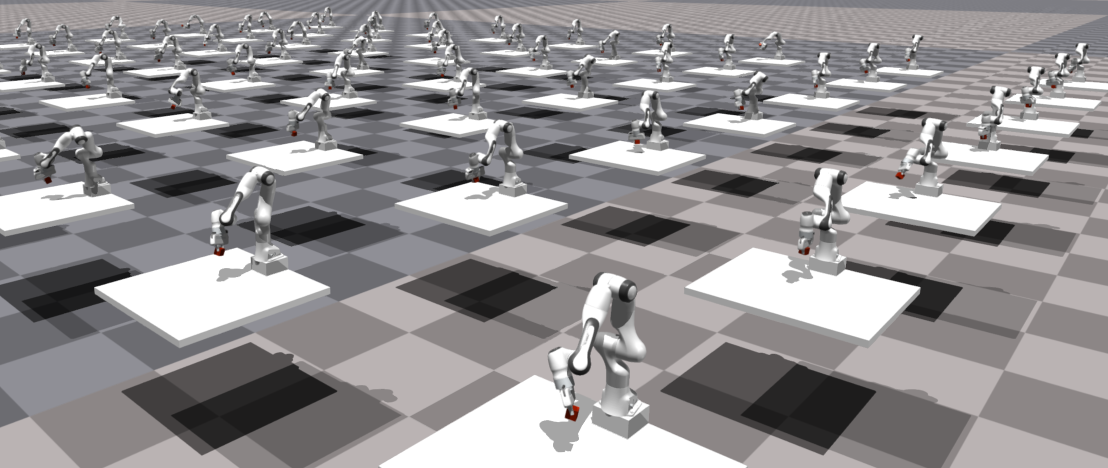}%
    \label{fig:isaac_two}}
  \hfill
{%
    \includegraphics[width=0.48\linewidth]{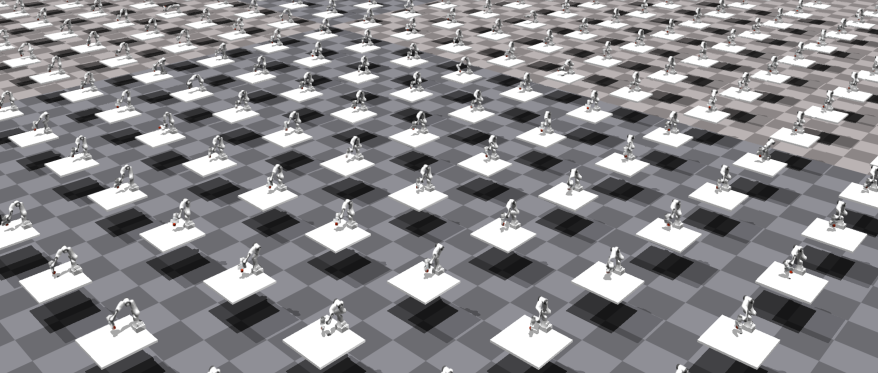}%
    \label{fig:isaac_one}}
  \caption{Visualization of large-scale parallel training in Isaac Gym. The left shows a close-up of agents interacting with targets, while the right displays a top-down view of 8,192 parallel environments. Each robot operates independently.}
  \label{fig:isaac_sim_envs}
\end{figure}

This section evaluates the robotic arm’s performance in target-reaching and grasping. The experiments employ a RL algorithm based on SNNs, incorporating CRL and dynamic reward modulation. As illustrated in Fig.~\ref{fig:isaac_sim_envs}, evaluations were conducted in the Isaac Gym simulator using 8,192 parallel environments to accelerate training and enhance sample efficiency.

\subsection{Experimental Setup, Training and Evaluation Procedure}

\subsubsection{Experimental Setup}

Experiments were conducted on the Isaac Gym platform using a customized training environment, running on a workstation equipped with an NVIDIA RTX 3090 GPU. The robotic platform consists of a 7-DoF Franka Emika Panda arm and a 2-DoF parallel gripper, forming a 9-DoF system. The manipulation task involves two stages: target approach and grasping. 

\subsubsection{Training Procedure}

To reduce computational complexity and systematically evaluate shallow architectures in high-dimensional control tasks, this study employs a SNN composed solely of input and output layers. Two key enhancements are integrated into the PPO framework: (1) a CRL strategy combined with dynamic reward modulation to promote stable convergence and enable progressive task acquisition; (2) the design of a delta observation vector that precisely captures the Euclidean distance between the end-effector and the target, thereby enhancing the spatial awareness of the state representation.

\subsubsection{Evaluation Protocol}

The results are averaged over 10 independent episodes, with cube positions randomized to assess generalization. Two baselines are used for comparison: one without CRL and delta observation, and one ANN with the same architecture and training settings. Evaluation metrics include task success rate, which is defined as the percentage of successful grasps per evaluation, and policy stability, which is assessed by monitoring reward convergence during training.

\subsection{Comparative Experiment}

\begin{figure}[htbp]
  \centering
{%
    \includegraphics[width=0.48\linewidth]{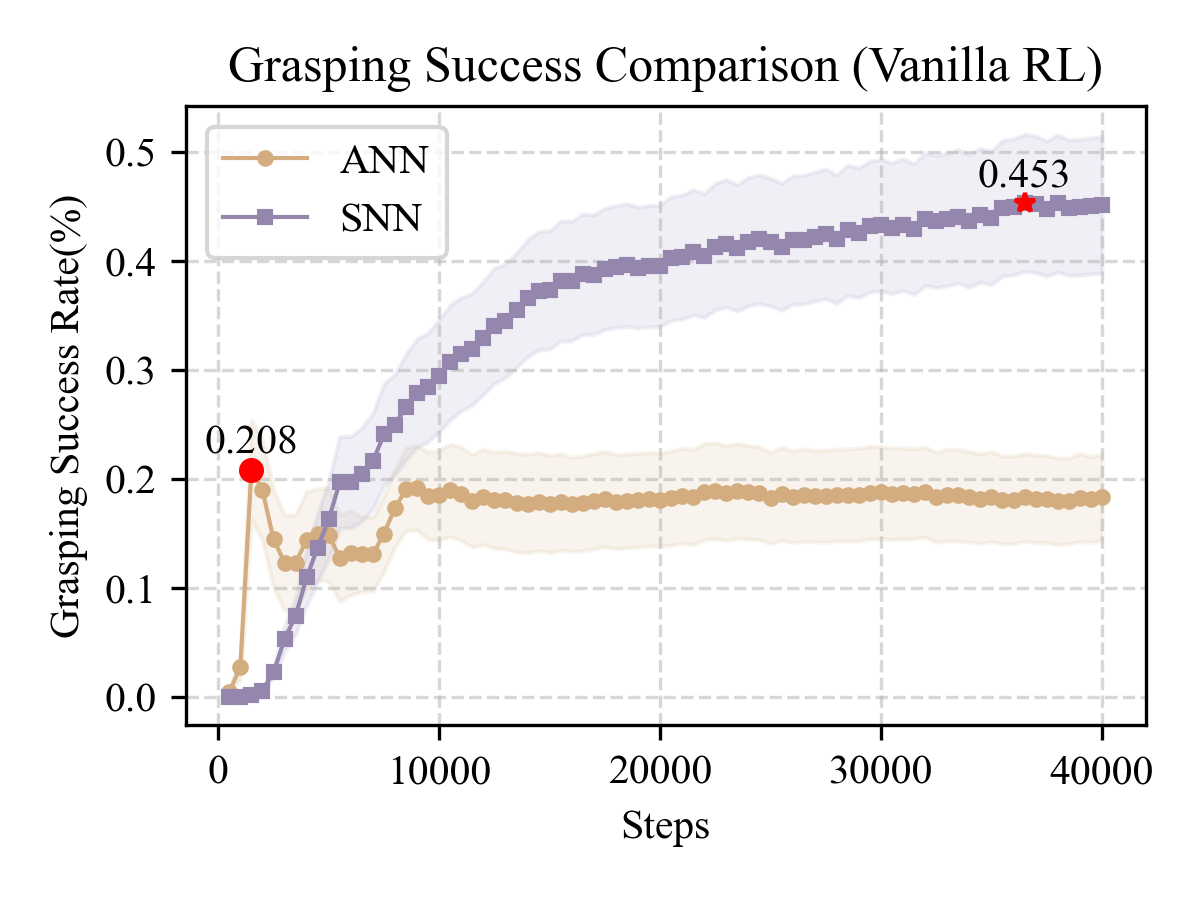}%
    \label{figurethree}}
  \hfill
{%
    \includegraphics[width=0.48\linewidth]{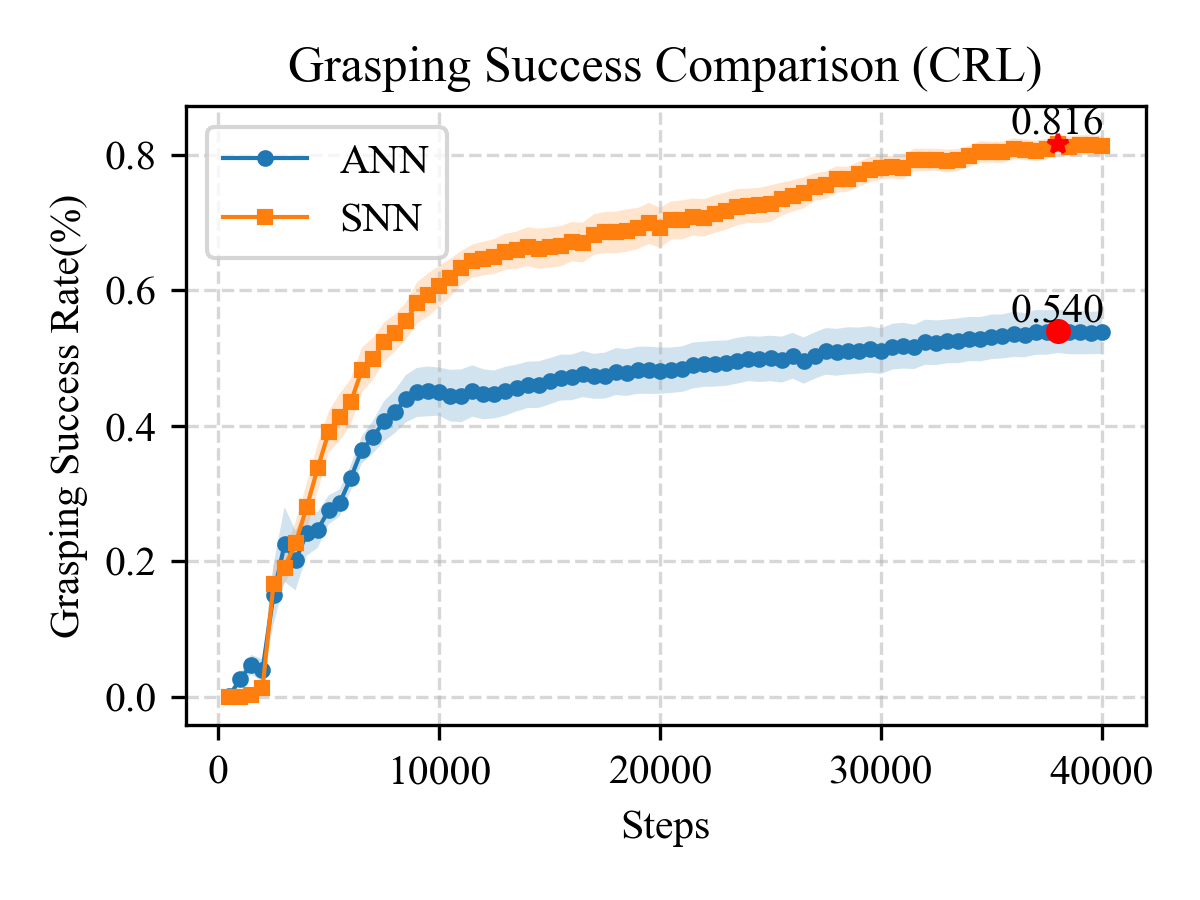}%
    \label{figurefour}}
    \caption{Grasping success rate of ANNs and SNNs under two different training frameworks. The left plot shows results under the Vanilla RL setting, while the right plot corresponds to the CRL framework with a dynamic reward strategy. Shaded regions indicate ±0.2 standard deviation. Red markers denote the peak success rate for each method.} 
  \label{figcatchsuccess}
\end{figure}

The effectiveness of the proposed shallow spiking actor-critic network was evaluated through comparative experiments involving two model types: SNNs and ANNs. Evaluation was carried out on a standard robotic manipulation task focused on object grasping. Two training configurations were considered: a baseline reinforcement learning setup (vanilla RL) and CRL with dynamic reward weighting. Results are summarized in Fig. \ref{figcatchsuccess}.

\subsubsection{Performance under vanilla RL}

In the grasping task, the SNN agent demonstrates a consistent and gradual improvement, reaching an average success rate of approximately 45\% by 40,000 steps. In contrast, the ANN agent shows an early rise followed by a plateau at around 15–19\%, with limited further gains. The results suggest that shallow SNNs can outperform comparable ANN architectures in both convergence and final performance under identical training settings.

Despite its minimal structure without hidden layers, the shallow SNN consistently outperforms its ANN counterpart under the standard PPO framework. This suggests that spike-based neural encoding enables more efficient information processing in action selection.

\subsubsection{Performance with CRL and dynamic reward weighting strategy}
 The curriculum strategy progressively increases task difficulty, encouraging the agent to acquire the full manipulation sequence—approaching and grasping in a stage-wise manner.

The proposed CRL framework significantly improved task performance across both SNNs and ANNs models. However, the SNN policy demonstrates faster and more stable gains. In the grasping task, the SNN surpasses a 70\% success rate within the first 20,000 steps, eventually stabilizing around 80\%. Although the ANN also improves, it plateaus at approximately 50\%.

\subsection{Reward Analysis}
\begin{figure}[htbp]
    \centering
    \includegraphics[width=0.5\textwidth]{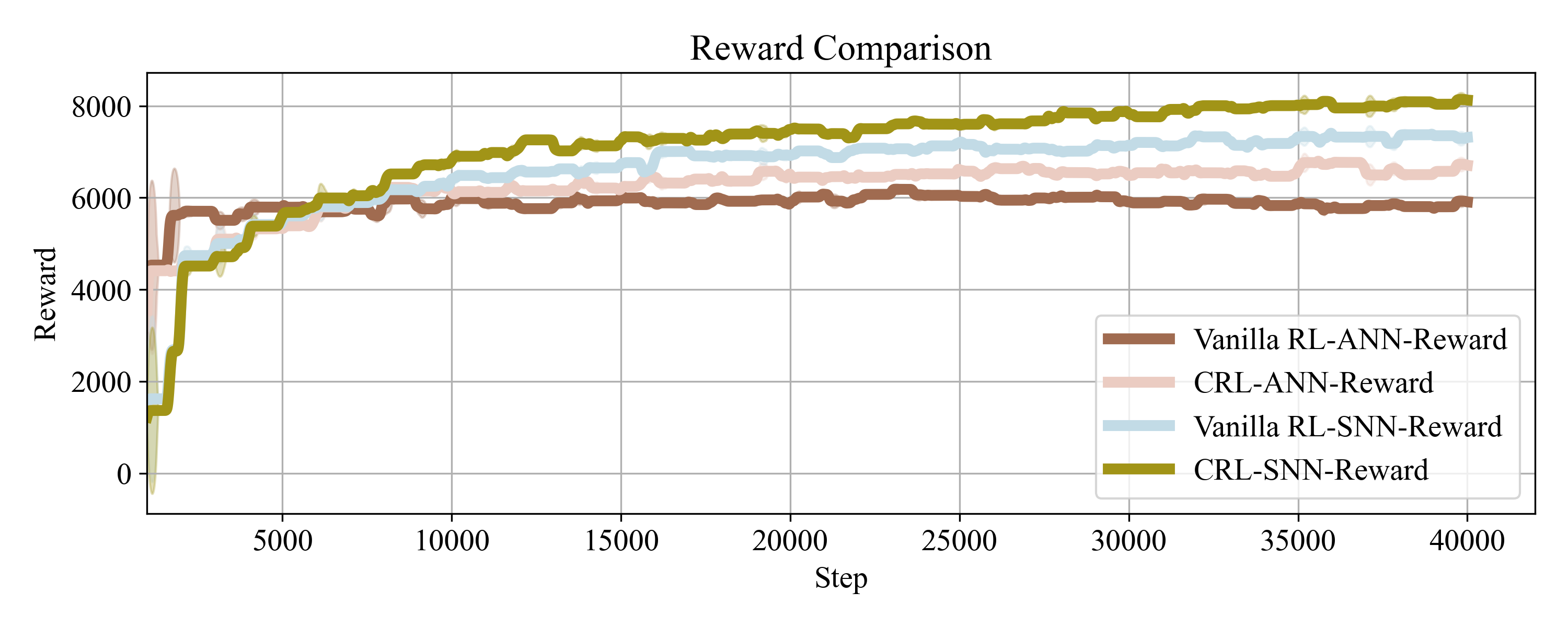}
    \caption{Reward trajectories under vanilla RL and CRL. The SNN trained with CRL shows faster reward improvement and achieves a higher, more stable final reward compared to other models. Each curve represents the average over 10 independent training runs, with smoothing applied to highlight overall trends.}
    \label{figurefive}
\end{figure}

\begin{figure}[htbp]
    \centering
    \includegraphics[width=0.4\textwidth]{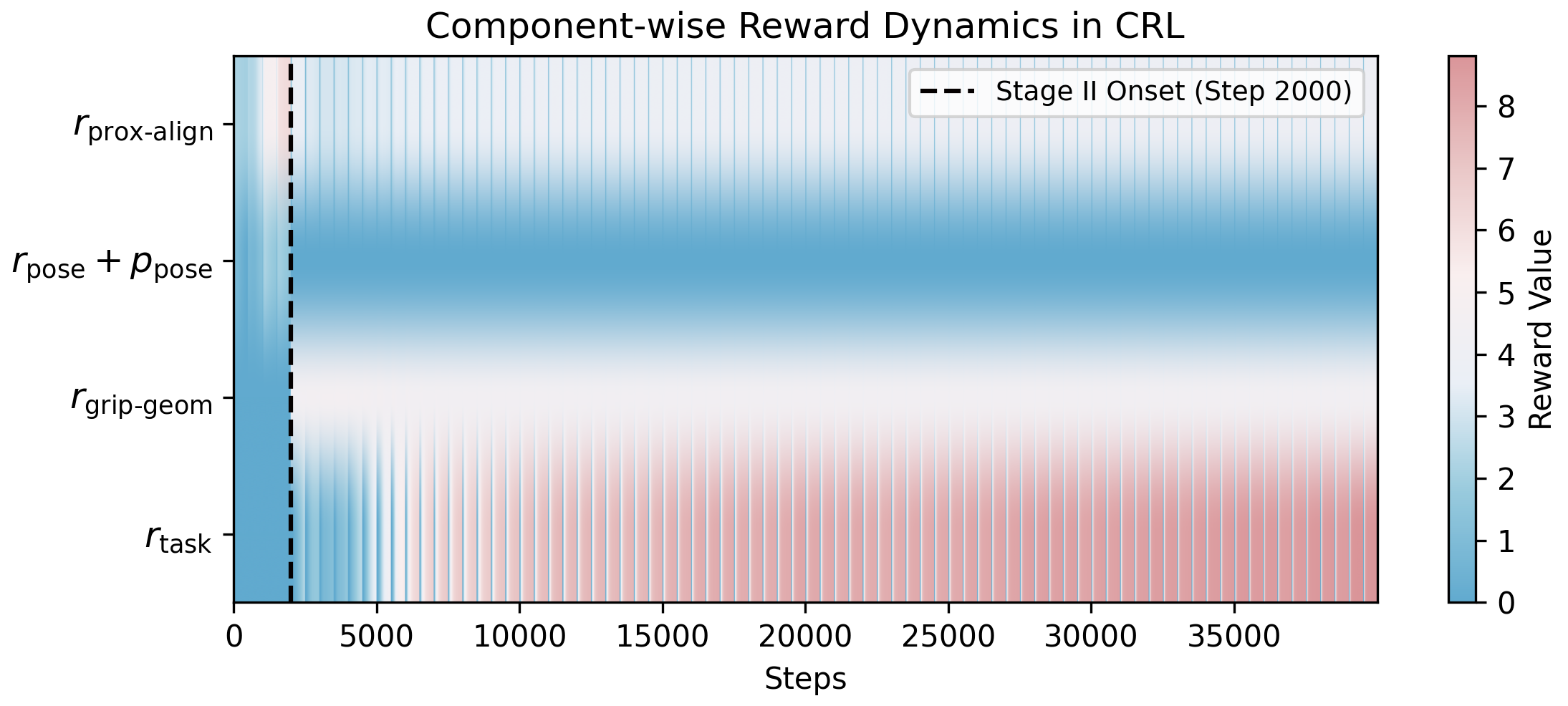}
    \caption{Evolution of individual reward components during training. The dashed line marks the transition to the second stage of training at step 2000, corresponding to the shift in reward emphasis under curriculum learning.}
    \label{figuresix}
\end{figure}

In addition to analyzing task success rates, the progression of cumulative episodic rewards was also tracked throughout the training process. This allows for a clearer understanding of the learning efficiency and convergence characteristics exhibited by the policies based on SNNs and ANNs. Fig. \ref{figurefive} presents the averaged results over ten independent trials, while Fig. \ref{figuresix} illustrates the fine-grained evolution of individual reward components in a single run, highlighting the shifting emphasis of the reward structure across different curriculum phases.

Under vanilla RL, both models show a sharp increase in rewards during the initial training phase, with the ANN slightly outperforming the SNN. However, after about 7,000 steps, the SNN continues to improve steadily and eventually surpasses the ANN. 

In the early phase of training with CRL, both the ANN and SNN models achieve substantial improvements, with the ANN policy initially attaining slightly higher rewards. However, beyond approximately 5,000 steps, the SNN continues to make steady gains, whereas the ANN reward curve begins to flatten. The finding suggests that although CRL contributes positively to both models, it provides a more substantial and sustained benefit to SNNs, particularly in terms of maximizing rewards and maintaining consistent learning performance.

\subsection{Energy Analysis}
The energy efficiency of the proposed fully SNN was assessed by comparing it with an ANN baseline on identical robotic manipulation tasks. Energy consumption was estimated using an analytical model, as detailed in Section~II-\textit{D 5)}, which takes into account both the types of arithmetic operations and the level of activation sparsity.

These values were substituted into Eq.(\ref{eq:snn_energy}) and Eq.(\ref{eq:ann_energy}) to estimate the theoretical total energy consumption of SNN and ANN models, respectively. Detailed results are presented in Table \ref{table_energy}.

\begin{table}[htbp]
\centering
\renewcommand{\arraystretch}{1.3}
\resizebox{\linewidth}{!}{  
\begin{tabular}{|c|c|c|c|c|c|c|c|c|}
\hline
\rowcolor{gray!25}
\rowcolor{gray!25}
\multicolumn{9}{|c|}{\textbf{CRL with Dynamic Reward Strategy}} \\
\hline
\textbf{Model} & ${r(r_{in})}$ & ${r_{mem}(r_{\text{out})}}$ & $B$ & $T$ & $N_0$ & $N_1$ & $N_2$ & $E_{\text{final(mJ)}}$ \\
\hline
SNN & 0.31 & 1 & 8192 & 500 & 18 & 256 & 7 & 38.49 \\
ANN & 1 & 0.48 & 8192 & 500 & 18 & 256 & 7 & 122.23 \\
\hline
\multicolumn{8}{|l|}{\textbf{Energy Saving}} & \textbf{68.51\%} \\
\hline
\end{tabular}
}  
\caption{Energy Consumption Comparison.}
\label{table_energy}
\end{table}

As shown in Table \ref{table_energy}, the proposed SNN achieves over 68\% energy savings compared to the ANN baseline under vanilla RL training. This efficiency arises from spike-driven sparsity and the use of low-cost accumulate operations across layers. 

\section{\label{exa}CONCLUSION}

This paper proposes a fully spiking actor-critic CRL framework to improve robotic manipulation under realistic physical constraints. The approach uses a simple SNN with only input and output layers, incorporates dynamic curriculum-based rewards, and expands the observation space to handle multi-stage tasks like reaching and grasping.

Experiments on the Isaac Gym platform show that despite its minimal network depth, the SNN-based agent achieves better training stability and higher task success rates compared to traditional ANNs. An energy modeling framework is also introduced to accurately estimate inference energy consumption, demonstrating that the proposed SNN significantly reduces energy use relative to ANN models.

Overall, this work offers a scalable and energy-efficient control method for robotic systems. Future work will focus on extending the framework to more manipulation tasks and deploying it on neuromorphic hardware.
\bibliographystyle{IEEEtran}
\bibliography{./conference_101719}
\end{document}